\documentclass[letterpaper, 10 pt, conference]{ieeeconf}  % Comment this line out if you need a4paper

\IEEEoverridecommandlockouts                              % This command is only needed if 
\usepackage{fancyhdr}
\fancypagestyle{withfooter}{
  
  \fancyfoot[C]{\footnotesize Accepted to the IEEE ICRA Workshop on Field Robotics 2024}
}
\usepackage{amsmath,amssymb,amsfonts}
\usepackage{algorithmic}
\usepackage{graphicx}
\usepackage{textcomp}
\usepackage{xcolor}

\usepackage{url}
\usepackage{hyperref}

\usepackage{xcolor,colortbl}
\definecolor{green}{rgb}{0.1,0.9,0.1}
\definecolor{purple}{rgb}{0.9,0.1,0.9}
\newcommand{\tickk}{$-$}
\newcommand{\tick}{\cellcolor{green}\tickk}
\newcommand{\cross}{$+$}
\newcommand{\cros}{\cellcolor{purple}\cross}

\usepackage{subfigure}
\usepackage{siunitx}
\def\BibTeX{{\rm B\kern-.05em{\sc i\kern-.025em b}\kern-.08em
    T\kern-.1667em\lower.7ex\hbox{E}\kern-.125emX}}

\usepackage[style=ieee]{biblatex}
\begin{document}

%\title{Unified Mapping Standards for Robotic Systems: Enhancing Interoperability and Efficiency Across Diverse Environments}
\title{\LARGE \bf Unified Map Handling for Robotic Systems: Enhancing Interoperability and Efficiency Across Diverse Environments}
% \title{Unified Map Handling for Robotic Systems: Enhancing Interoperability, Efficiency, and Deployment Across Diverse Environments}
%\title{Unified Map Handling for Robotic Systems: Enhancing Interoperability and Efficiency Across Diverse Environments and Heterogeneous Fleets}

\author{James R. Heselden$^{1,2}$ and Gautham P. Das$^{1,3}$% <-this % stops a space
\thanks{*This work was not supported by any organization}% <-this % stops a space
\thanks{$^{1}$Lincoln Institute of Agri-Food Technology, University of Lincoln, UK}
\thanks{$^{2}$\{jheselden@lincoln.ac.uk\}, \{0000-0001-6494-4981\}}%
\thanks{$^{3}$\{gdas@lincoln.ac.uk\}, \{0000-0001-5351-9533\}}%
}

\maketitle

\thispagestyle{withfooter}
\pagestyle{withfooter}

\begin{abstract}
Mapping is a time-consuming process for deploying robotic systems to new environments. The handling of maps is also risk-adverse when not managed effectively. We propose here, a standardised approach to handling such maps in a manner which focuses on the information contained wherein such as global location, object positions, topology, and occupancy. As part of this approach, associated management scripts are able to assist with generation of maps both through direct and indirect information restructuring, and with template and procedural generation of missing data. These approaches are able to, when combined, improve the handling of maps to enable more efficient deployments and higher interoperability between platforms. Alongside this, a collection of sample datasets of fully-mapped environments are included covering areas such as agriculture, urban roadways, and indoor environments.
\end{abstract}

% \begin{IEEEkeywords}
% map generation, information conversion, environment representation
% \end{IEEEkeywords}

\section{Introduction}

% \subsection{Motivation}

% what is a map?
A map can be described as a diagrammatic representation of the ontology of a space \cite{oxford_definition}. This encodes the spatial arrangement of physical features. There are many forms this can take such as: grid maps showing occupancy, to topological maps showing traversability, and world maps showing explicit object positions. Maps traditionally show a variety of physical features, however, modern map formats are used to encode more virtual and meta information.

% why one map is not enough
For robotics systems, maps are a necessity as such systems are unable to sense and perceive all the information about a space in real-time. Constructed maps are used here to represent a subset of information in a manner accessible for the robot to read and respond to in real-time. Depending on the purpose of the map, different information can be encoded and stored.
% how we get metric maps
In deployments to a new location, significant time may be taken to create all the necessary maps to allow it to function. 

Many outdoor services can rely on GNSS-based navigation such as used in \cite{clearpathTopo} and indoor navigation using topologies-based navigation such as in \cite{anyboticsTopo}. For agricultural environments, much of the environment can be viewed from satellite imagery, so generating maps can be a trivial task by hand. However, for more complex maps, necessary for more sophisticated activities by precision agricultural machinery, work must be completed to extract the information necessary for mapping \cite{katikaridis2022uav,santos2020occupancy}. In contrast, for indoor environments such as warehouses, office buildings, and shopping centres, occupancy mapping is most often completed with SLAM approaches \cite{macenski2021slam}.

% why multi-agent systems should share maps
Consistency between environmental representation is a necessity in the management and coordination of robotic fleets. Guaranteeing maps are both up-to-date and consistent across robots is key to ensuring interoperability and efficiency in dynamic environments. Given a map is a representation of an environment, if the environment is changed with new static obstacles in place or new routes opened up, all robots should be aware of these changes, and have a shared common knowledge. Through the sharing of this knowledge, robots can be more adaptive to changes, increasing responsiveness and reducing downtime.

% how maps stored?
Currently, standard practices for handling maps in single-robot deployments is often ill-defined. This leads to developers storing map files within the same file-structure as the rest of the codebase. In contrast, with larger fleet deployments the maps are often stored in a centralised database, on dedicated infrastructure such as an edge-server. In these instances, the maps are distributed and updated live. When maps are stored within other code used for running robot-specific tasks, this can cause problems in consistency as mentioned above. When spaces are being used by multiple independent robotic systems, this type of map handling can lead to multiple maps being updated, with different understandings of the environment. This can lead to problems such as miscommunication between robots where locations are registered under different names, or collisions with obstacles when maps are not up to date with the latest environmental changes.

The main contributions of this work are the open, flexible standardisation of environment representation; and the demonstration of how such an approach can both enhance interoperability and improve deployment efficiency through enabling pipeline development for automating cascaded deployments. Additionally, the paper contributes an open dataset of sample environments in a variety of environment structures.

The remainder of this document is as follows. Section \ref{sec:information} details the types of information contained within map files, and how various commonly-used maps are structured. Section \ref{sec:reformatting} gives an overview of the types of conversion approaches which have been developed under this work, along with an introduction into procedural generation approaches. Section \ref{sec:template} walks through the template structure used for the work, and the public datasets being provided. Finally Section \ref{sec:conclusion} summarises the findings and contributions of this work.

\section{Environmental Information}
\label{sec:information}
\subsection{Information Types}

In this section, we overview some types of information available from an environment and the manners in which this information is stored and used.

        \subsubsection{Location}
Location information refers to the location of the space on a wider scale, in particular, this is the type of information used to transform the location of a robot through GNSS and apply them to a location within the space's metric positioning.

        \subsubsection{Objects}
Objects are the items which exist within the space, their location, properties and interactions within the space. This type of information is used to give an understanding of objects which can be interacted with such as the location of a door in a room.

        \subsubsection{Occupancy}
Occupancy is the information about object permanence, that being how the world is filled by obstacles and where those obstacles couple against the potential movements of the robot. The position of walls in a 2d space or manner in which a table occupies a room and the space in which the table does not occupy in a 3D space.

% For example, to plan a route through an environment, a robot must have information regarding obstacles in the environment and regions it is able to move through. For most purposes, the necessary information can be encoded as an occupancy grid in which total vertical occupancy is encoded into a binary image file wherein black and white pixels show occupied and unoccupied spaces \cite{occupancy_grid}. If the robot is airborne, total vertical occupancy misses regions such as under tables, in which the drone may be able to traverse. In this case, an octomap may be more appropriate as more information can be encoded into the map in a 3D representation \cite{octo_map}. 

        \subsubsection{Topology}
The topology of a space detail the connection of regions within the space, how rooms connect to one another, and how an agent is able to pass through such regions.

% To further supplement a system, further maps may be used to give the robot more information about the environment. A topological map can be used to further inform a robot about which routes through the environment it is able to use without it having to perform intense calculations on a occupancy map or octomap \cite{topological_map}.

% how we get topological maps
% The construction of topological maps and network maps, is often done through navigating a space and dropping way-points, or built by hand with GUI applications \cite{topo_rviz_tools, gui_topo_tool}, with some approaches in outdoor environments using autonomous generation from satellite imagery \cite{topo_satellite_generation}. Automated approaches such as these have shown reductions in map generation time and faster deployment to new environments, however are each focused on a specific domain and are thus not as generic and transferable.

\subsection{Information Encoding}

Below are a variety of file types which encode different types of information, both directly and indirectly. Although this is not an exhaustive list, most of the variety of information required for autonomous robot operations is available in these. Table \ref{table:information_types} summarises the types of information stored in various map types, showing how some information types can be extracted directly from the files, and how some can be generated directly from the same files. Notable, is how often, topology can be extracted indirectly, while in contrast location cannot be extracted without this information being supplied.

\begin{table}[ht]
    \caption{Information stored within maps formats, information available in raw form (shown with \tickk) and indirect information available without use of complementary files (shown with \cross).}
    \label{table:information_types}
    \centering
    
    \begin{tabular}{r|cccc|c}
        \textbf{Map Type} & 
                     \textbf{Location} & 
                             \textbf{Objects} & 
                                     \textbf{Occupancy} & 
                                             \textbf{Topology} & 
                                                     \textbf{Ref} \\ \hline
             Datum   & \tick &       &       &       & [1] \\
         OSM Graph   & \tick & \cros &       & \tick & [2] \\
           KML/KMZ   & \tick & \cros &       & \tick & [2] \\
           OpenRMF   &       & \tick & \cros & \tick & [3] \\
         Satellite   &       & \tick & \cros & \cros & [4] \\
            Gazebo   &       & \tick & \tick & \cros & [5] \\
          Grid Map   &       &       & \tick & \cros & [6] \\
          Octo Map   &       &       & \tick & \cros & [7] \\
          NavGraph   &       &       &       & \tick & [8] \\
          TopoMap2   &       &       &       & \tick & [9] \\
        
    \end{tabular}
\end{table}

% \newpage
    
    \subsubsection{KML File}
Keyhole Markup Language (KML) is an international standard for use in rendering geographic information in an Earth-browser \cite{ogc_kml}. This file format contains three levels of detail, Points, Polylines and Polygons. Points are single locations with just a latitude, longitude and elevation, while Polylines are a series of coordinates, and Polygons are Polylines which connect at the end. Zipped KMZ files can optionally contain, alongside the KML file, complementary files such as icons, images, overlays, and COLLADA 3D models for further detailing in rendering information.
    
    \subsubsection{Datum File}
Datum data used for localisation of world coordinates to global coordinates \cite{ros_navsat_transform_node}. This file is formatted in YAML. It contains standard properties such as a latitude, longitude of the centre coordinate in the local coordinate space exists within the global coordinate frame. Additionally, it includes a GNSS fence for defining the boundaries of the region of interest, gmapping map size, mapviz origin points, and navsat transform points, all for use with standard ROS systems for utilisation of GNSS-based localisation and mapping.
    
    \subsubsection{Satellite Image}
Image file captured from a satellite or drone, or birds-eye view of an outdoor space, or conversely, a floor plan of an indoor space. Such images can contain top-down object information and topological data, however such formats require decoding for extraction.
    
    \subsubsection{Occupancy Grid}
Occupancy file used to manage occupancy in 2D space, used with the ROS navigation stack \cite{ros_move_base, ros_nav2}. Information is stored in two files, an image file in PGM or PNG and a properties file stored as YAML. The image file shows a grey-scale or binary image in which occupation is encoded for each cell. In this file, a scale is used to determine occupancy with white showing free space, and black showing occupied space. The accompanying properties file is used to reference (i) the map image file,  (ii) the relative position of the image to the world, (iii) the size which each cell represents in the real world, (iv) the point on the scale at which a cell should be regarded as occupied, and (v) the position on the scale at which a cell should be regarded as free.

    \subsubsection{Octo Map}
Occupancy file used to render occupancy efficiently in three-dimensional space \cite{hornung2013octomap}. Information is stored in binary Octree format which stores space occupation under a tree where each level of the tree stores the region subdivided into a 2x2x2 grid in which each sub-division is either fully empty, fully occupied, or partially occupied containing another sub-tree. Each voxel can optionally encode RGB colour information. Limiting query-depth can enable higher performance by reducing the level of detail.
    
    \subsubsection{Open-Street Map}
Open-Street Maps content is used for a variety of systems in government and commercial operations \cite{osmcartographyTopo}. Information is stored in XML. This map type details geographic data such as buildings, roads, and city boundaries, including properties such as names, tower types, speed limits, road surfaces, population estimates, and a large assortment of properties for different types of markings. The types of entries are categorised into (i) Nodes which are single points such as buildings, (ii) Ways which are collections of nodes connected in a sequence, and (iii) Relations, which are collections of Ways with a shared relationship.
    
    \subsubsection{NavGraph}
Topological file used for basic detailing of networks of permissible routes within an environment under the FawkesRobotics simulator \cite{fawkesTopo}. Data is stored in YAML format with information structured in three sections. In (i), default properties detail the tolerances of how closely the robot should follow a path. Nodes (ii) define the list of nodes which each have a position, name, and any special typing which require special behaviours. Connections (iii) include a list of node name pairs to show paths between nodes.
    
    \subsubsection{Topological Map}
Topological file used to enable high-level navigation through environments. Data is stored in YAML format and detailed in \cite{topological_map}. Information contained within includes local-space positions of nodes and connections to neighbouring nodes. Each edge detail restrictions of which robots are permitted awareness of the edge, and each node details a boundary and tolerance of which the agent may use to localise.
    
    \subsubsection{OpenRMF Map}
Simulation world file for use in the OpenRMF ROS2 Traffic Simulator \cite{openrmfTopo}. Data is stored in YAML format and contained information is broken down into Crowd Sim which details dynamic agents such as robots and humans to include into the simulations. Under Levels, it also details: (i) static obstacles on each floor of the building, such as models, floors, and walls, (ii) interactive objects such as doors, and (iii) lanes which robots may pass along, and human lanes for use by humans.

    \subsubsection{Gazebo World}
Simulation world file for use in the Gazebo Simulator \cite{gazebo_sim}. Data is in SDF, an XML-flavour format specifically designed for simulation definitions. Data includes: (i) static object positions like walls, shelves, and tables, (ii) dynamic obstacles such as animals or people, (iii) intractable objects such as doors and elevators, and (iv) environmental properties such as lighting, weather, and gravity.

\begin{figure}[ht]
    \centering
    \subfigure[Satellite view of region]{
        \label{fig:map_empty}
        \includegraphics[width=.2\textwidth]{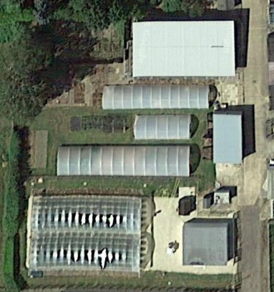}
    }
    \subfigure[Datum-based GNSS fence]{
        \label{fig:map_datum}
        \includegraphics[width=.2\textwidth]{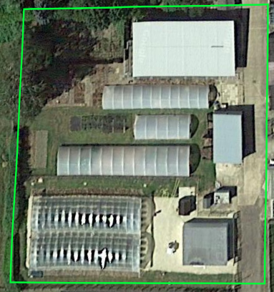}
    }
    \subfigure[Occupancy grid]{
        \label{fig:map_occup}
        \includegraphics[width=.2\textwidth]{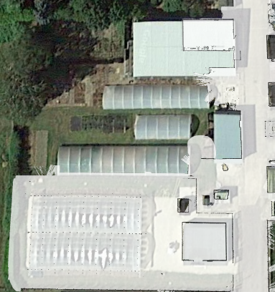}
    }
    \subfigure[Topological map]{
        \label{fig:map_topo}
        \includegraphics[width=.2\textwidth]{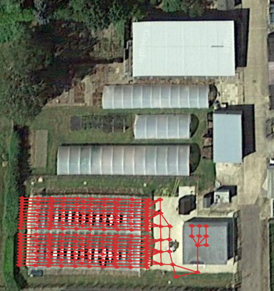}
    }
    \caption{Collection of maps rendered using Google Earth using KML.}
\end{figure}

\section{Information Reformatting}
\label{sec:reformatting}
\subsection{Applications of Conversions}
With understanding the type of information stored within such map files, we can find approaches to extract this information reuse in other formats. For example, by extracting the path information from an OpenRMF file, we can generate a topological map. We can further supplement available information with the use of API calls, i.e. we can request a satellite image from an API when given a datum fence. A satellite image can in turn, be used to extract object information for use in generating a simulation. Allowing the construction of pipelines for cascading the generation of environments.

\begin{figure}[ht]
    \centering
    \includegraphics[width=0.9\linewidth]{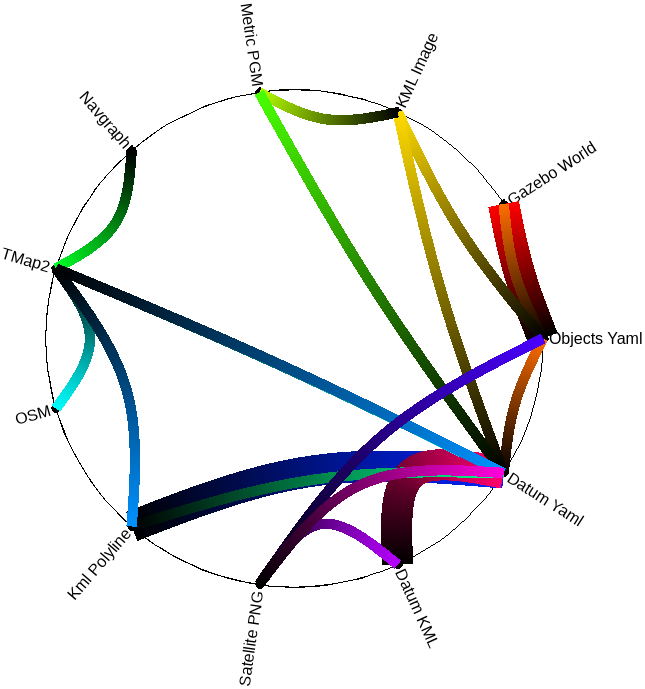}
    \caption{Chord Diagram detailing examples of conversions between different standard map types. Each line indicates a conversion from the black end of the line to the destination file. Each conversion process is indicated by a different destination colour with multiple lines being used together.}
    \label{fig:chord_diagram}
\end{figure}

Fig.\ref{fig:chord_diagram} shows the conversion approaches which have been developed thus far as part of the open-source converters stack in our $environment\_common$ \footnote{\url{https://github.com/LCAS/environment_common}} structured as a ROS2 package. Each colour details a different converter, with multiple lines ending in the same colour indicating a combination of those files to construct the destination file type. Due to the limited sources of information on location data, the datum files have been used extensively for localising to the world space, a necessity for KML generation.
    
Detailed below, are some of the approaches considered for extracting information from existing files for then combining them into new formats.

In its fundamental form, a topology is a collection of points and a list of how these points connect together. This information is used to represent the connections in a space and how a robot can move through the space. This type of information, as detailed in Table.\ref{table:information_types}, can be generated implicitly through occupancy information. As shown in \cite{santos2019path}, this can be done through morphological skeletonisation in which permissible space in an occupancy grid is repeatedly thinned to leave only single lines and connecting points.

Occupancy, showing how things within a region reserve space, can be determined in several ways. In \cite{katikaridis2022uav}, occupancy is implied through the projection of trees onto an occupancy grid. While in \cite{santos2020occupancy}, occupancy is determined from satellite image segmentation. For the generation of occupancy maps from object-based maps, approaches such as \cite{gazebo_slicer} have been developed to slice through simulated worlds at a given height. In \cite{devi2022feature}, the authors develop an approach to segment satellite imagery, with effective results in classifying paths and roads. This type of segmentation could be effective to further extract topology information with approaches such as in \cite{santos2019path}.

In the identification of objects, many works have been developed for effective image segmentation \cite{kirillov2023segment}. These, when combined with effective labelling provided by such works as YOLOv7 \cite{wang2023yolov7}, and deployed on a drone, can be an efficient way to collect object information from the real world. Approaches such as YOLOv7 \cite{wang2023yolov7} could also be applied to satellite images to gather this type of information, albeit in lower quality, from satellite images prior to visiting the field.

Some information on objects can be obtained from occupancy information, especially when minor context is given such as what objects could possibly be within a given space. For instance, if it is known that the deployment is in a forest, and an occupancy map is offered, then it could be reasonable to assume obstacles which appear to have a circular occupancy may in fact be trees, and in an indoor office, that flat edges to the occupancy could be walls.

\subsection{Types of Procedural Generation}

When restructuring environmental data for new purposes, not all information can be obtained through conversions of existing files and use of API calls. Some pieces of information may be missing, as they were never collected in the first place. To fill in these pieces of missing data, and to fabricate entire environments from scratch, procedural generation can be used. 

For instance, the conversion of a NavGraph map, to an Topological Map, there is information about node positions and connections which is retained, however information on which navigation approach to use for each connection simply does not exist. This information could be supplemented by hand with generation of new files for integration, however with the use of template data to auto-fill the missing fields, this can be skipped.

When template data is not enough, and more realistic and reasonable data must be generated, procedural generation can be used instead. Detailed below are a few examples of the types of procedural generation which can be applied to auto-fill missing data while keeping environments realistic.

    \subsubsection{Wave-Function Collapse}
Inspired by quantum theory, wave-function collapse works by creating and collapsing states based on compatibility with a template structure. In practice, a template is provided describing a likely sample of a full generation, and the system iteratively generates parts of the environment and destroys parts which are not in the template sample.
    
    \subsubsection{Cellular Automata}
Approaches in Cellular Automata work to generate complex patterns from simple rules. Specifically, they work by simulating rule-based interactions of cells on a grid where the state of each cell is dependent on the state changes of neighbouring cells.
   
    \subsubsection{Gradient Noise}
Gradient Noise is often used for generating texture maps. It works by applying an algorithm over a grid to construct natural-looking patterns. This approach is used commonly in the forms of Perlin Noise which uses a square-grid structure, and Simplex Noise which uses a triangle-based grid structure.

    \subsubsection{Procedural Modelling}
Procedural Modelling is used for constructing types of models with defined architectural traits. L-Systems are a well-known approach of this which is primarily used for procedural generation of plants and foliage. It works by utilising structural rules about how individual components relate to one another in order to build up simulated architectural models. ArcGIS CityEngine by Esri is another well-known approach which is used for urban planning and design, which generates city models using GIS data, or through rule-based generation.

\section{Environment Template}\label{sec:template}
The $environment\_template$ \footnote{\url{https://github.com/LCAS/environment_template}} uses a fixed file structure in which files are categorised into their primary data category. This ensures proper organisation and a tidier workspace. At the root of this hierarchy is a source file named $environment.sh$. Each file included within the file structure should also have a standard export included into this source file. This is to enable flexible file referencing wherein regardless of the specific file locations within the hierarchy, users can always identify the file they need through sourcing the statically positioned source file.

The environment template hierarchy is contained within a ROS2 package format. This is to enable simple utility, within ROS2 infrastructure projects, and to enable compatibility with the map conversion scripts which utilise ROS2 for tidier file management and execution. It is important to stress that the environment template is not dependent on ROS2 for execution, the template of choice can be downloaded to any location on the device so long as the source file is referenced appropriately.

Example datasets of environments have been made public, created under this template. These are available on the environment template repository, and are detailed further on the repository wiki. A short breakdown of each is detailed below:

\subsection{Agricultural Dataset}
\begin{figure}[ht]
    \centering
    \subfigure[Full map of Riseholme Park Farm]{
        \label{fig:dataset_agri}
        \includegraphics[width=.42\textwidth]{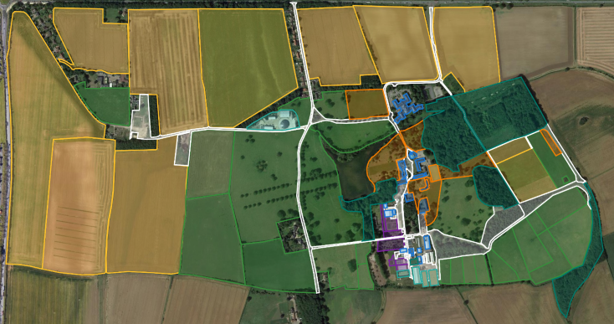}
    }
    \subfigure[Central region of Riseholme Park Farm zoomed in]{
        \label{fig:dataset_agri_zoom}
        \includegraphics[width=.42\textwidth]{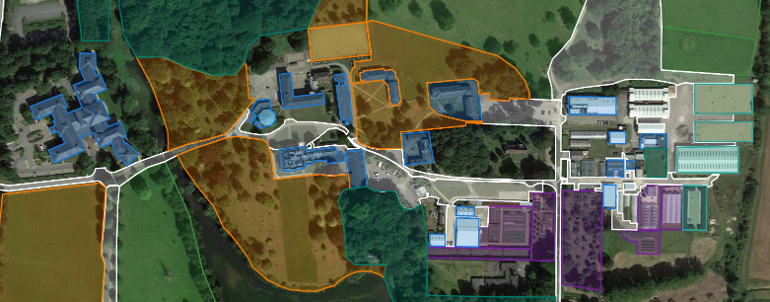}
    }
    \caption{Regions included with the agricultural environment dataset, rendered with Google Earth. Colours of polygons indicate the type of space, with yellow showing horticultural fields, green indicating pastoral field, dark green showing forest, orange showing parkland, blue showing office buildings, cyan showing animal housing, and purple showing grower plots for specialist crops. White regions indicating connecting paths and roads across the farm.}
    \label{fig:agri}
\end{figure}

The agricultural environments dataset is generated over Riseholme Park Farm (Fig. \ref{fig:agri}). The dataset, consists of a collection of fully and partially mapped regions, covering horticultural fields, pastoral fields, grower plots, research buildings, parkland, woodlands livestock buildings, along with connecting roads and footpaths. Further mapping is ongoing, with new maps being generated alongside new conversion processes.

\subsection{Urban Roadway Dataset}
\begin{figure}[ht]
    \centering
    \subfigure[Locations across Asia used for urban roadway environments.]{
        \label{fig:dataset_road_asia}
        \includegraphics[width=.2\textwidth]{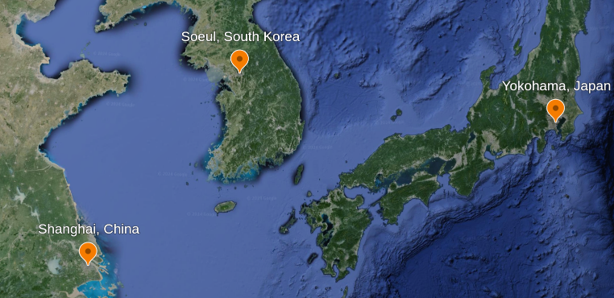}
    }
    \subfigure[Locations across Europe used for urban roadway environments.]{
        \label{fig:dataset_road_europe}
        \includegraphics[width=.2\textwidth]{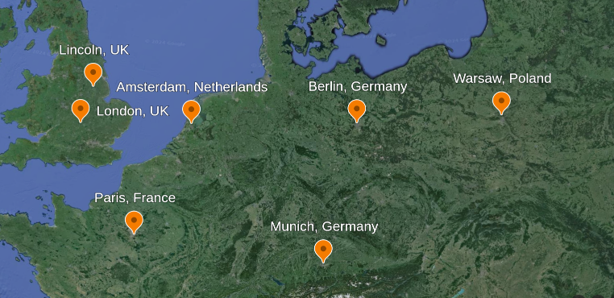}
    }
    \subfigure[Rendering in KML, of OSM data over Ixtapa converted into a topological map.]{
        \label{fig:dataset_road_osm_kml}
        \includegraphics[width=.42\textwidth]{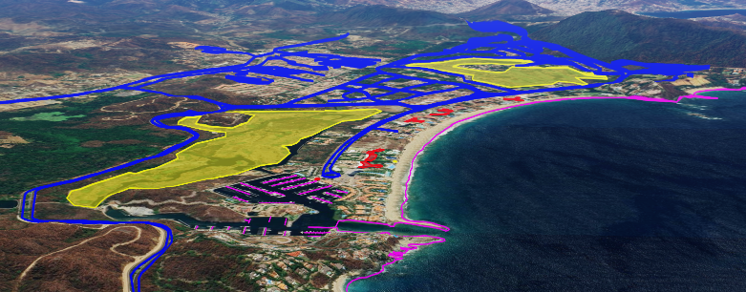}
    }
    \subfigure[Locations across North America used for urban roadway environments.]{
        \label{fig:dataset_road_america}
        \includegraphics[width=.42\textwidth]{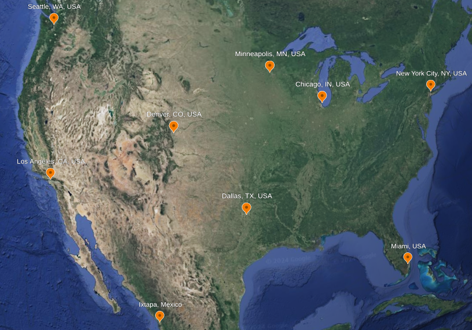}
    }
    \caption{Locations across the globe in the Urban Roadway Environment dataset.}
    \label{fig:road}
\end{figure}

The urban roadway environments dataset consists of a collection of environments generated from the open-street map dataset. It includes a collection of roadway maps from cities in Asia (Fig. \ref{fig:dataset_road_asia} - e.g. Tokyo, Soeul, Singapore), Europe (Fig. \ref{fig:dataset_road_europe} - e.g. London, Madrid, Berlin) and North America (Fig. \ref{fig:dataset_road_america} - e.g. New York, Los Angeles). Road network in Ixtapa, converted from OSM to Topological map format and embedded on Satellite imagery in KML is shown in Fig. \ref{fig:dataset_road_osm_kml}.

\subsection{Indoor Dataset}
\begin{figure}[ht]
    \centering
    
    \subfigure[\SI{10}{\metre\squared} space]{
        \label{fig:dataset_ware1}
        \includegraphics[width=.2\textwidth]{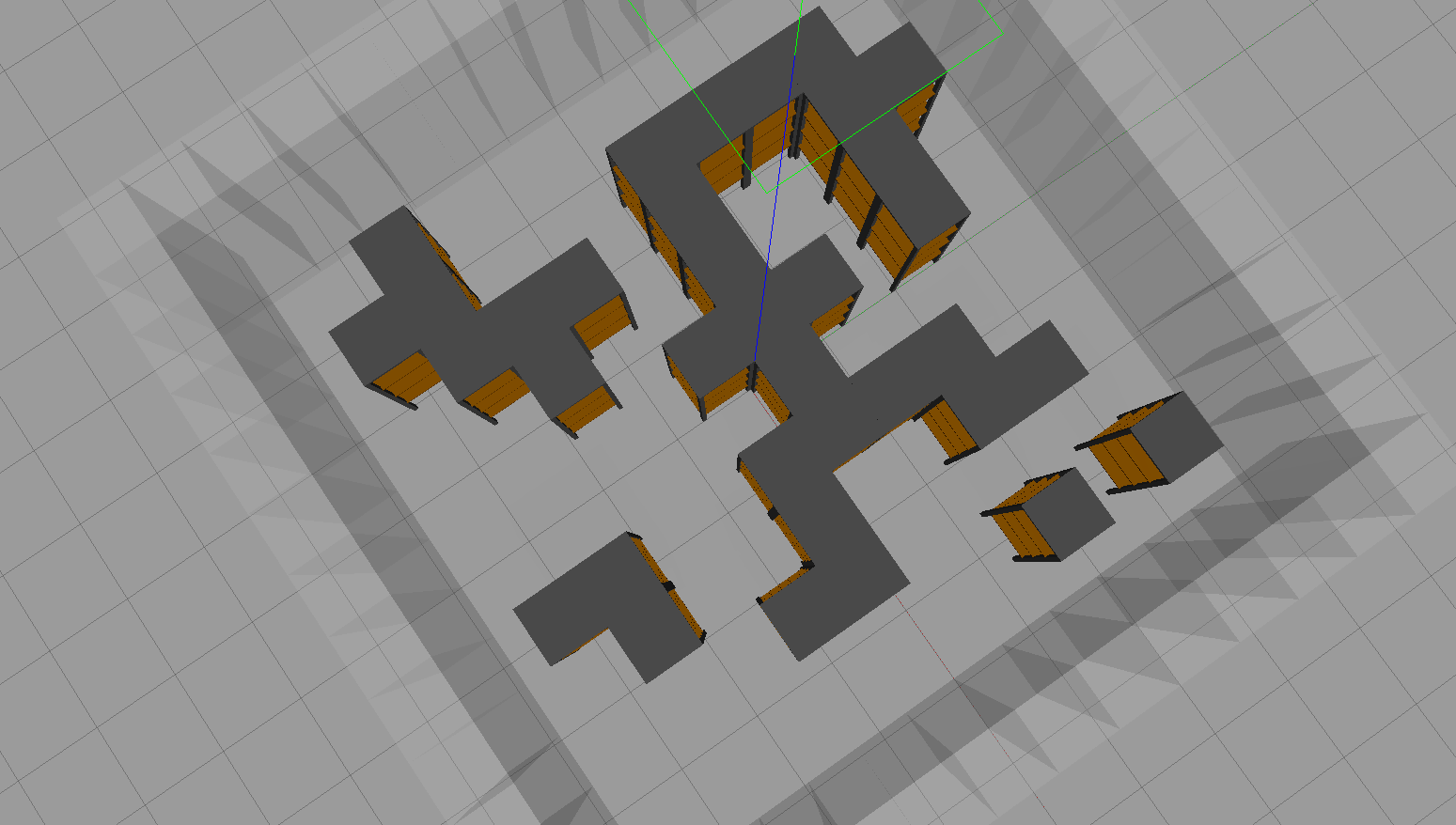}
    }
    \subfigure[\SI{15}{\metre\squared} space]{
        \label{fig:dataset_ware2}
        \includegraphics[width=.2\textwidth]{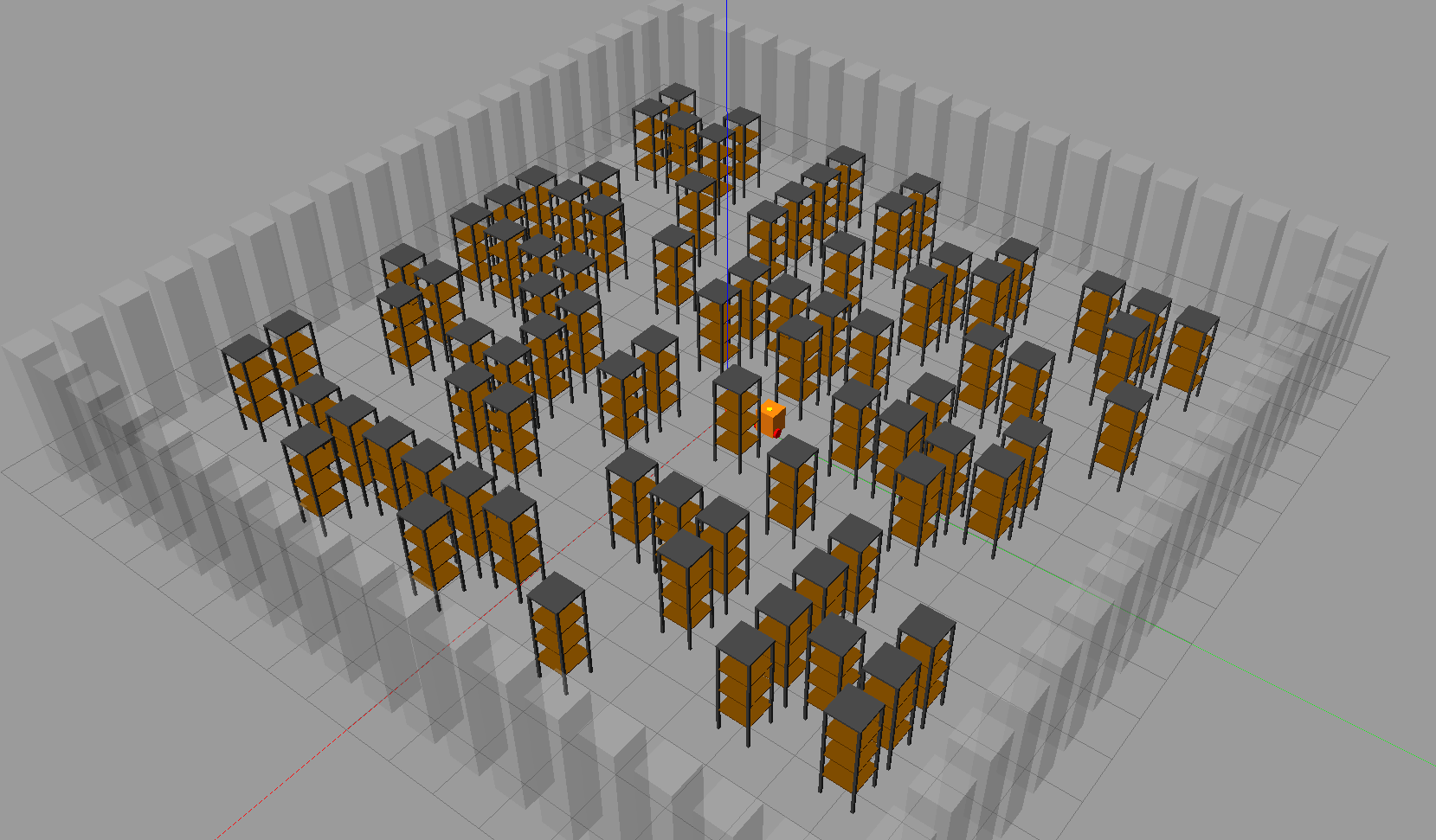}
    }
    \caption{Examples warehouse environments generated procedurally using wave-function collapse. Both examples had been supplied with a \SI{5}{\metre\squared} sample template detailing an example of an how an arrangement of warehouse shelves and free space may appear on a 2D grid. (a) shows the result of a \SI{10}{\metre\squared} generated regions, and (b) shows the result of a \SI{15}{\metre\squared} generated region.}
    \label{fig:indoor}
\end{figure}

The indoor environment dataset consists of a collection of indoor environments some converted from existing sources, and some generated procedurally using wave-function collapse. It includes environments such as offices, warehouses, airports and hospitals (Fig. \ref{fig:indoor}).

\section{Conclusion}\label{sec:conclusion}
Mapping is still a major hurdle for the rapid deployment of technologies into new environments. This challenge is multi-fold when considering the long-term deployment of heterogeneous robotic fleets that may use different map formats for their navigation planning and execution. 

In this work, we have shown the potential benefits that employing map-handling standards can have for improving this rapid deployment. Alongside this, we have detailed the necessity of sharing maps to ensure consistency and interoperability between robots running different code-bases. Collating all information related to an environment in one independent package would also alleviate the challenges of scattered environment information and updates, especially in large-scale deployments. We have also highlighted the potential of environment generation pipelines, with direct and indirect information extraction, complemented with the use of API calls and procedural generation for determining missing data. 

However, this work is not complete without wider usage in the robotics community and identifying much more efficient ways to ensure interoperability and information completion. Towards this, we have opened a large dataset of environments covering agricultural spaces, urban road networks, and warehouses, which are highly effective for comparative experimental analysis. We have also open-sourced the standards, with the intent for this dataset to be expanded further through community development.


\section*{Acknowledgements}
This work was supported by AgriFoRwArdS CDT, under the Engineering and Physical Sciences Research Council [EP/S023917/1]. This work was completed in association with the Agri-OpenCore project under Innovate UK grant 10041179.

\printbibliography

\end{document}